\documentclass[conference]{IEEEtran}
\IEEEoverridecommandlockouts
\usepackage{cite}
\usepackage{amsmath,amssymb,amsfonts}
\usepackage{algorithm}
\usepackage{algpseudocode}
\usepackage{multicol}
\usepackage{graphicx}
\usepackage{textcomp}
\usepackage{xcolor}
\def\BibTeX{{\rm B\kern-.05em{\sc i\kern-.025em b}\kern-.08em
    T\kern-.1667em\lower.7ex\hbox{E}\kern-.125emX}}
\begin{document}

\title{Investigating the Effectiveness of a Socratic Chain-of-Thoughts
(SocraCoT) Reasoning Method for Task Planning in Robotics, A Case Study\\
{\footnotesize \textsuperscript{*}}
}

\author{
\IEEEauthorblockN{Veronica Augustina Bot} \\
Stanford University \\
\texttt{botva@stanford.edu} 
\and
\IEEEauthorblockN{Zheyuan Xu} \\
IEEE Member \\
\texttt{cx1014@uw.edu} \\
\thanks{Repository Link on Github: \textcolor{blue}{https://github.com/CharlesXu1124/Socracot}}
}


\maketitle

\begin{abstract}
Large language models (LLMs) have demonstrated unprecedented capability in reasoning with natural language. Coupled with this development is the emergence of embodied AI in robotics. Despite showing promise for verbal and written reasoning tasks, it remains unknown whether LLMs are capable of navigating complex spatial tasks with physical actions in the real world. To this end, it is of interest to investigate applying LLMs to robotics in zero-shot learning scenarios, and in the absence of fine-tuning - a feat which could significantly improve human-robot interaction, alleviate compute cost, and eliminate low-level programming tasks associated with robot tasks. 

To explore this question, we apply GPT-4(Omni) with a simulated \textit{Tiago} robot in Webots engine for an object search task. We evaluate the effectiveness of three reasoning strategies based on Chain-of-Thought (CoT) sub-task list generation with the Socratic method (SocraCoT) (in order of increasing rigor): (1) Non-CoT/Non-SocraCoT, (2) CoT only, and (3) SocraCoT. Performance was measured in terms of the proportion of tasks successfully completed and execution time (N = 20). Our preliminary results show that when combined with chain-of-thought reasoning, the Socratic method can be used for code generation for robotic tasks that require spatial awareness. In extension of this finding, we propose EVINCE-LoC; a modified EVINCE method that could further enhance performance in highly complex and or dynamic testing scenarios. 
\end{abstract}



\section{Introduction}

Unlike text-only applications, robotic systems require a deep understanding of real-world
physics, environment context, and the ability to perform physical actions through mastery of actuation systems and low-level controllers. A generative
robotics model must have robust commonsense knowledge and a sophisticated world model,
with the ability to execute commands in ways that are physically possible in the real world \cite{lamport94}. These challenges fall beyond the original scope of language models, as they must not only understand the meaning of a given text, but also translate the intent into a logical sequence of actions with generated code.

Chain-of-thought (CoT) reasoning is a promising method of enhancing LLM output for complex arithmetic, symbolic, and commonsense tasks \cite{Yao3}. Inspired by the step-by-step thinking ability of humans, CoT has been applied to language models to solve multi-step reasoning problems [3]. This in-context-learning is a training-free paradigm that has been shown to significantly boost model performance and data efficiency across NLP benchmarks in few-shot scenarios [3]. Given the complexity of spatial tasks, multi-modal inputs, and the need for modular generated code, CoT is thought to be a promising tool for integrating language models and robotics [4].

Although known to enhance LLM reasoning, CoT has shortcomings in sub-task planning and code generation [5, 6]. Prior work has shown that compared to CoT alone, multi-agent debate outperforms single model baselines in reasoning, factuality, and question-answering tasks. [6] In this regard, supplementing CoT with structured multi-agent adversarial debate could potentially provide more reliable action sequences, particularly for complex spatial tasks whose performance is historically limited. 

Recent work [7] extends the chain-of-thought approach by adding a diverse set of reasoning paths and performing majority voting among independent LLMs. The Socratic method employs deductive, inductive, and abductive reasoning to ensure consistency and accuracy of inference. The Socratic method deals with all aspects of critical thinking, including definition, clarification, and cross-examination [7]. This comprehensive approach to reasoning can lead to improved output quality and consistency [7]. Of interest in this work is whether or not the  Socratic method can enhance the logic of generated sub-task lists and coding sequence(s).

\begin{figure*}[!htbp]
    \centering
    \includegraphics[width=1\linewidth]{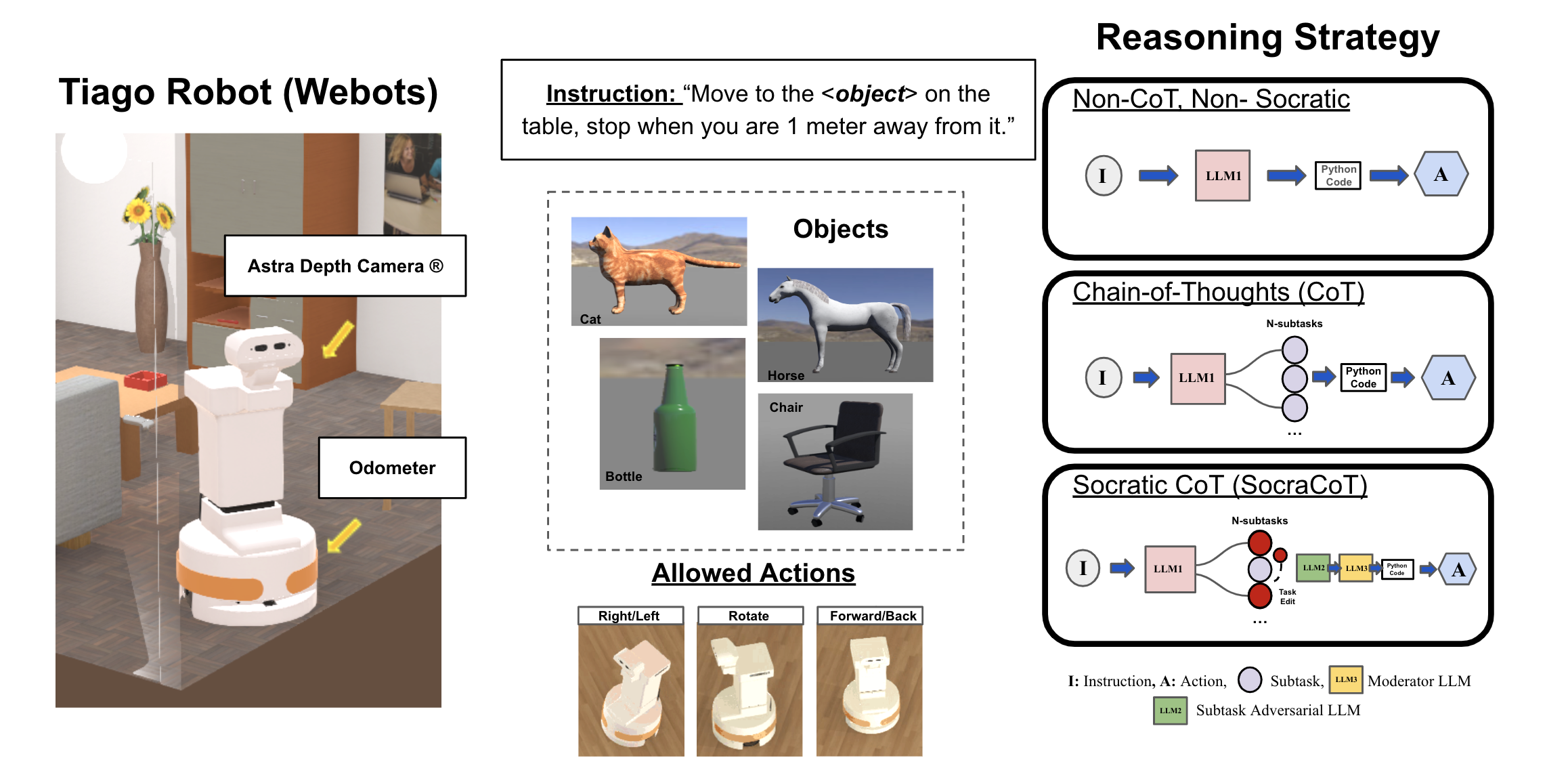}
    \caption{Experimental setup for evaluating the effectiveness of a Socratic CoT reasoning strategy when applied to robotics in a simulated environment. Leftmost picture shows the simulated Tiago robot, which is equipped with an odometer placed on its chassis, two differential wheels and one Astra Depth camera placed on its head, which can output both RGB and depth images. The robot also has a Hokuyo Lidar on its chassis for obstacle avoidance. The middle picture details the objects used for testing and the possible actions for the robot. The rightmost picture shows the 3 scenarios we tested and benchmarked against, the first one being vanilla setup without any chain-of-thought nor Socratic participation, the second one includes chain-of-thought strategy, and the third one employs Socratic method on both chain-of-thought strategy and task code generation for completing the designated task described in natural language.}
    \label{fig:setup}
\end{figure*}

In this preliminary study, our objectives are two-fold. First, demonstrate the application of LLMs
to robotics for a simulated real-world task. Given the aforementioned limitations of CoT alone, we then evaluate the effectiveness
of a modified Socratic Chain-of-Thought (CoT) reasoning strategy in terms of proportion of tasks successfully completed and time to execute.

\section{Methodology}

\subsection{Simulation Environment}

Webots (version R2023b) is an open source multi-platform desktop application that supports modeling, programming and simulating robot interactions with applied objects and force regimes. Robots have different locomotion schemes (wheeled robots, legged robots, or flying robots) and are equipped with a number of sensor and actuator devices including distance sensors, drive wheels, cameras, motors, touch sensors, emitters, and or receivers. Robots are programmed by the user via the controller console. In this study, we utilized a simulated office environment in Webots ROS2 package to connected to the back-end powered by GPT-4(Omni). For consistency, we use GPT-4(Omni) in all the LLM queries in the context of this experiment. A simulated Tiago Lite robot is used in the simulation as described in \textit{Figure \ref{fig:setup}}. The robot is tasked with moving to an object in a predetermined array of positions in the office environment and stop within a specified distance. The details of the experiment setup can also be found in \textit{Figure \ref{fig:setup}}.

\begin{figure*}[!htbp]
    \centering
    \includegraphics[width=1\linewidth]{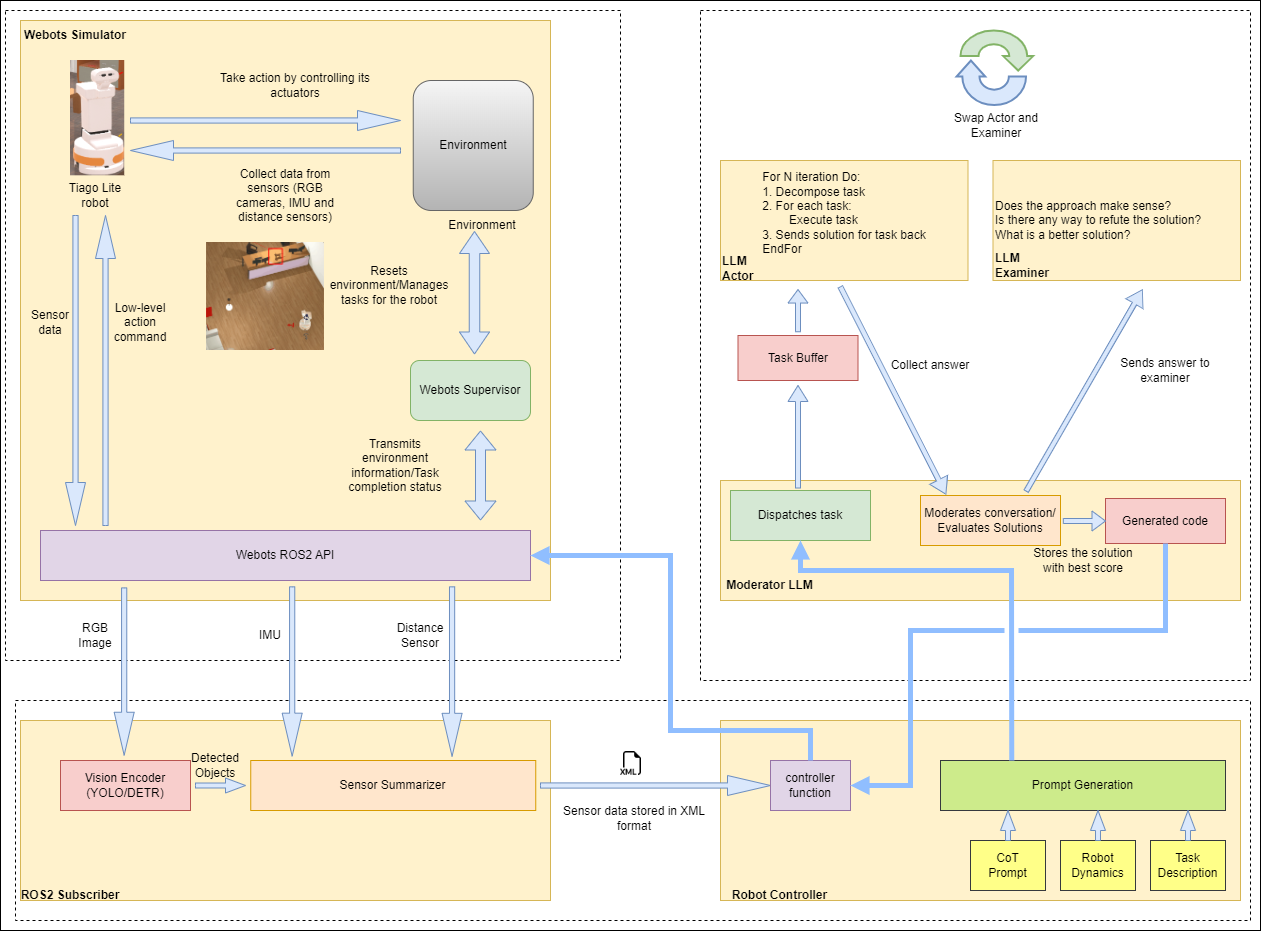}
    \caption{Proposed SocraCoT framework in high-level view, it consists of an actor LLM and an examiner LLM, whose roles get interchanged as they are "debating" over many iterations. The bottom component is responsible for constructing prompt, making queries to LLM API endpoint, collecting responses, and postprocessing the responses to control the robot. This module also subscribes to the sensor data published by the Webots ROS2 bridge, and makes decisions by calling the "move-robot" function generated by the LLM. The upper right component is responsible for LLM query and debate, by firstly asking the LLM (actor) to generate a list of subtasks, then by employing Socratic method, makes another LLM (critic) query to generate a revised list of subtasks. Afterwards, queries are made to LLM (moderator) for generating the code for the incomplete function "move-robot". The code generated is then reviewed and improved to ensure its robustness. The generated code is then plugged into the running Python script and executed in real time.}
    \label{fig:architecture}
\end{figure*}

\section{Task Description}
The high-level task we assigned to the robot is move to certain distance from a target object. The task can be further broken down into the following components:
\begin{itemize}
    \item \textit{Object Identification:} the Tiago robot must identify a target object in the office environment. The environment is supplied with non-target objects common to that setting. We implemented four separate objects of varying size and complexity including a green bottle, white horse, orange cat, and black leather chair. 
    \item \textit{Starting Position}: The Tiago robot is placed at a starting position within 10 meters of the target object facing away from it. We cycle the robot through a set of starting positions for each object search task and test performance over a series of runs. 
    \item \textit{Control Commands}: The robot is capable of turning right, left, advancing forward, reversing, and stopping. The robot can rotate in place (turn radius of 0 meters) or on a range of arc radii (0 meters to 15 meters). The command used to control the robot is in the format of Twist message in ROS2 geometry\_twist/msgs, which consists of velocity waypoint commands in linear and angular components. 
    \item \textit{Stopping Position}: The Tiago robot must utilize its Astra depth camera, explore the surroundings, move on a path of any geometry toward the target object and stop within a specified distance from the object. 
\end{itemize}
The LLM is given some contextual information about the robot and its environment, as well as its control interface script in Python. The LLM needs to complete the incomplete function  "move\_robot", to guide the robot and complete the task. The LLM is not finetuned, nor does it receive example outputs beyond the aforementioned content.

\section{Architecture}
There are 3 major components in the implementation as shown in \textit{Figure \ref{fig:architecture}}. 
\subsection{Simulation environment and robot}
The first component is the simulated robot and its environment in Webots as discussed above. The robot interacts with the environment, moves around by controlling its differential wheels after receiving velocity control waypoint commands, and collects sensor data from its odometer and depth camera. The depth camera outputs both RGB and depth image in 640$\times$480 pixels, and the odometer outputs robot positional and velocity data in ROS2 nav\_msgs/Odometry message. 
\subsection{Controller and prompting}
The second component is responsible for control and prompt generation, and acts as a middleman between the simulated environment and the LLM API endpoint. The controller node subscribes to both the depth camera and odometer ROS2 publisher, and calls controller function move\_robot every 10 frames, which translates to 0.1 seconds in simulation time.

\subsection{LLM query and response postprocessing}
The third component is responsible for coordinating debate between LLM models and collecting useful responses to control the robot as detailed below:
\begin{itemize}
    \item $\text{LLM}_A$ is firstly tasked to break down the high level task into a list of doable subtasks $L_A$
    \item Apply the Socratic method to ask $\text{LLM}_B$ to evaluate the response $L_A$, and re-generate a revised list of tasks $L_B$.
    \item Provide robot control interface script in Python as prompt, and the $L_B$, ask $\text{LLM}_C$ to generate code $C_c$ incomplete controller function move\_robot, and collect response.
    \item Apply the Socratic method to ask $\text{LLM}_D$ to revise the generated code, and reflect whether it suffices to complete the designated task. Meanwhile, the robot is asked to think of corner cases in which $C_c$ fails to cover, and re-generate the modified code $C_d$, which is then plugged into the original control interface script and executed every time sensor data gets updated. 
\end{itemize}

\section{Prompt Design}
\begin{figure}[!htbp]
    \centering
    \includegraphics[width=1\linewidth]{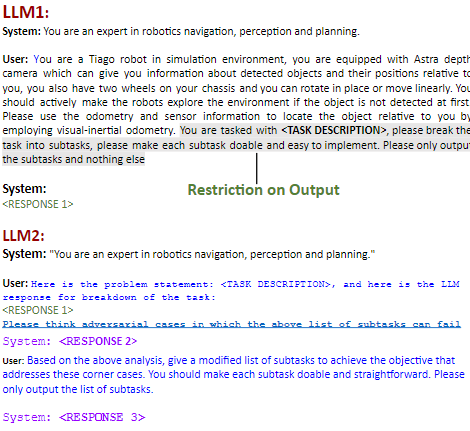}
    \caption{Prompt template used for breaking down the high-level task.}
    \label{fig:prompt_cot}
\end{figure}

\begin{figure}[!htbp]
    \centering
    \includegraphics[width=1\linewidth]{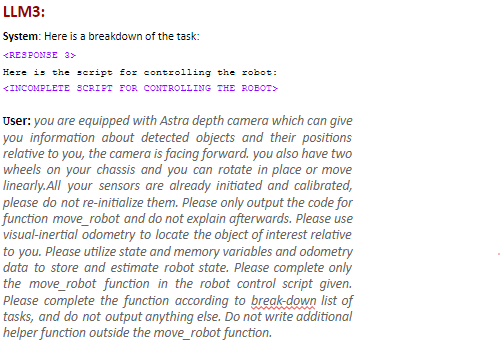}
    \caption{Prompt template used for generating the code for completing the robot controller function.}
    \label{fig:code_gen}
\end{figure}

\begin{figure}[!htbp]
    \centering
    \includegraphics[width=1\linewidth]{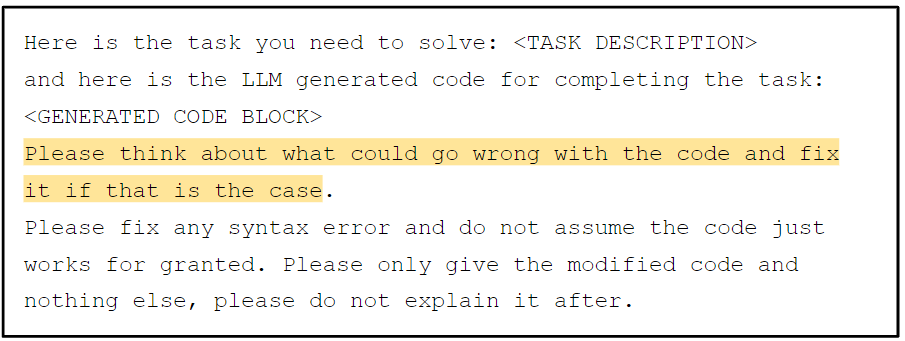}
    \caption{Prompt template used for revising the code through the method of counterfactual reasoning.}
    \label{fig:code_gen2}
\end{figure}

The high-level task given to the robot is described in natural language, as in \textit{Figure \ref{fig:architecture}}, and is in the the following format: 
“Move to the $<$object$>$ on the table, stop when you are $<$N$>$ meter away from it."
In which object and N are user-defined. Zero-shot chain-of-thought is then used for generating a list of doable subtasks as described in \textit{Figure \ref{fig:prompt_cot}}. The response is then collected and another query to LLM endpoint is performed to generate a modified list of subtasks. In this process, the LLM is encouraged to think of adversarial cases in which the previous list of tasks may fail. As described in \cite{EY2023}, this corresponds to the method of dialectic, which explores the corner cases and makes the search for answer diversified. After collecting the response, a third query to LLM is required to generate the incomplete function move\_robot, which utilizes updated sensor data and sends the robot velocity waypoints control command by constructing a ROS2 Twist message, and publishes it, as described in \textit{Figure \ref{fig:code_gen}}. Here, the method of Mieutics is employed, as it proposes generated code based on a list of subtasks to be followed. A final query to another LLM is then performed by making it act as a strict code reviewer that actively seeks for syntax errors, reflects upon the logic and supplies revised lines for completing the move\_robot function, by employing the method of counterfactual reasoning, as described in \textit{Figure \ref{fig:code_gen2}}.

\section{Evaluation}

During each experimental run (for an object and starting position pair) the robot is awarded 0-1 point for task completion. Scoring is performed by two independent human observers to ensure fairness. A consensus score is reported. A \textbf{score of 0 }is awarded if the robot fails to find the target object, produces an incomplete path, or prematurely stops due to an error in the generated code. A \textbf{score of 0.5 } is awarded if the object is identified and path is traversed, but the robot fails to stop in the designated 1 meter radius from the target object. A \textbf{score of 1 } is awarded if the robot completes the task as described.

\begin{figure}[!htbp]
    \centering
    \includegraphics[width=1\linewidth]{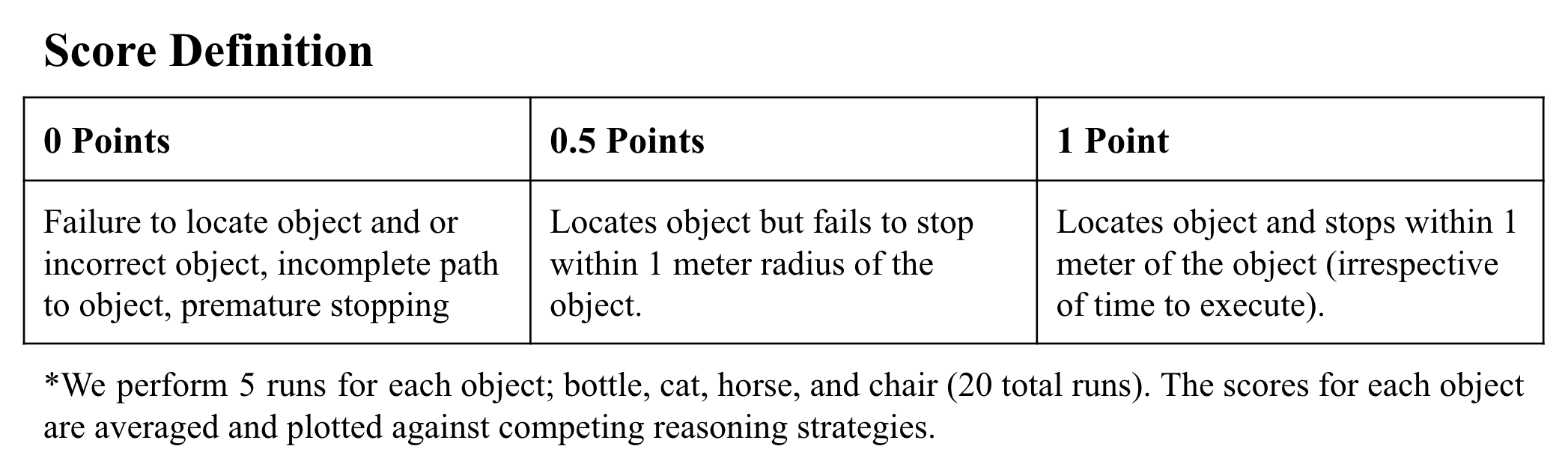}
    \caption{Scoring method used to evaluate task completion outcomes}
    \label{fig:scoring_method}
\end{figure}

Of note, we do not apply a time limit for execution. We allow the experimental run to continue until the task is completed or aborted due to failures described above. If the robot displays no motion, we prematurely stop the run and assign a score of 0, as shown in \textit{Figure \ref{fig:scoring_method}}. \textbf{A total of 20 runs were executed; 5 runs per object (bottle, cat, horse, chair)}. The total cumulative score was computed for each object and plotted for Non-CoT/Non-SocraCoT, CoT, and SocraCoT reasoning strategies. The maximum possible score is 5 - denoting task completion for all 5 runs for each object. The time associated with execution was recorded and averaged as described above (SD error bars represented). 

\section{Results}
\begin{figure}
    \centering
    \includegraphics[width=1\linewidth]{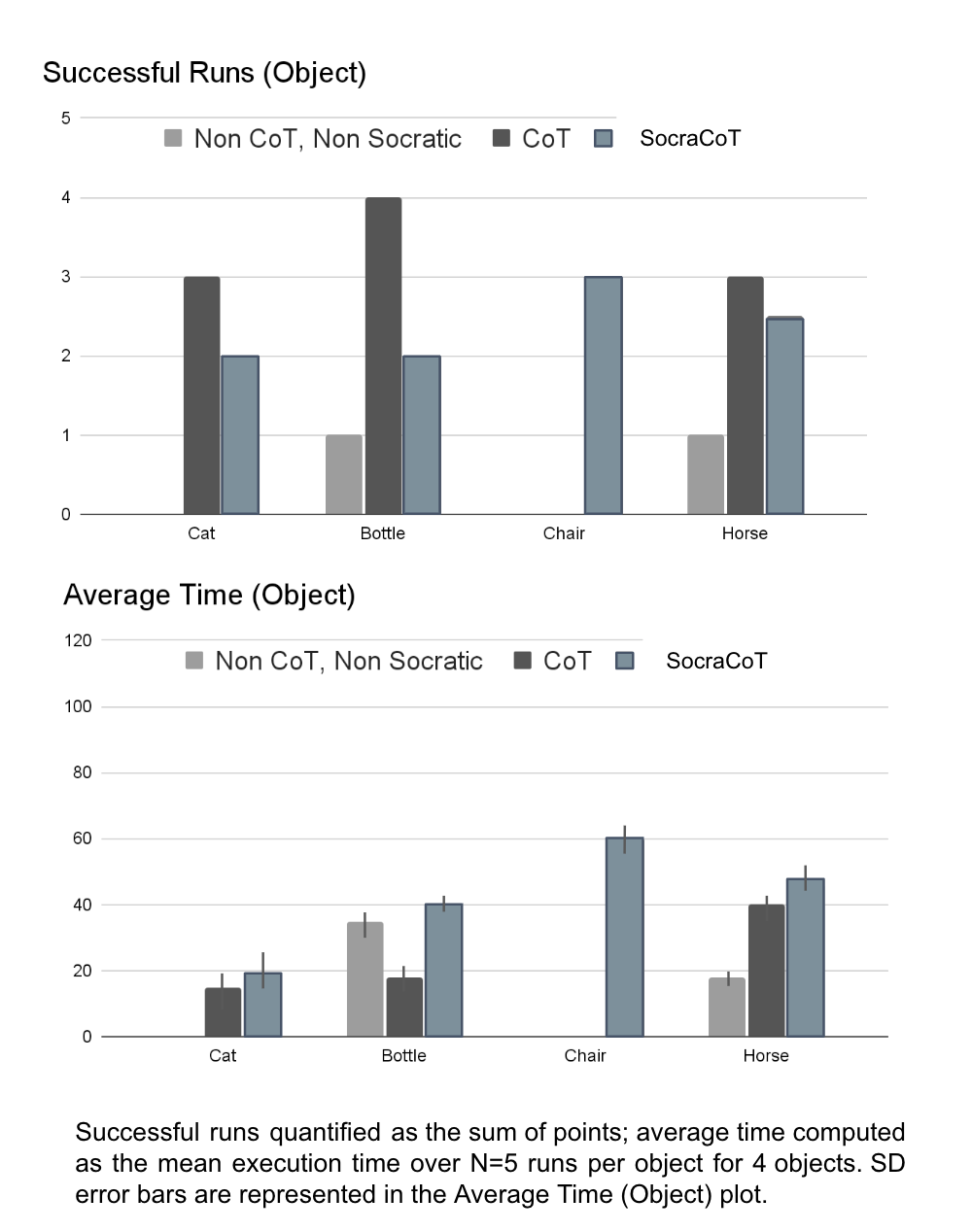}
    \caption{Task completion outcomes scored by cumulative score for successful runs, and time to execute (seconds). Scoring was performed by two observers. Consensus scores are represented.} 
    \label{fig:outcome}
\end{figure}

In this study, we sought to explore the effectiveness of a Socratic chain-of-thoughts reasoning method (SocraCoT) for completing a spatial task with robotics. Compared to CoT alone, Socratic reasoning shows potential in strengthening logic in generated sub-task lists, as plotted in \textit{Figure \ref{fig:outcome}}. This is especially useful for robotics, wherein a generated chain-of-thoughts must not only decompose a task, but also ensure it can be translated to a logical coding sequence in the real world. To our knowledge, this is the first time a study has been performed that combines the Socratic method and chain-of-thoughts reasoning to generate code for robotic task planning and execution.

Both CoT and SocraCoT reasoning strategies revealed improved task outcomes with higher successful run count compared to the non-CoT/non-SocraCoT approach. Of the failed runs for non-CoT/non-SocraCoT, a majority of errors were attributed to code failure or incorrect object identification. Both CoT and SocraCoT revealed less frequent code errors and instead failed due to poor object tracking and distance measurement (missing the 1m stopping radius, thus being awarded 0.5 points). Although seemingly more effective for task completion, there was a significant execution time and token cost associated with CoT and SocraCoT. 

\begin{figure}
    \centering
    \includegraphics[width=1\linewidth]{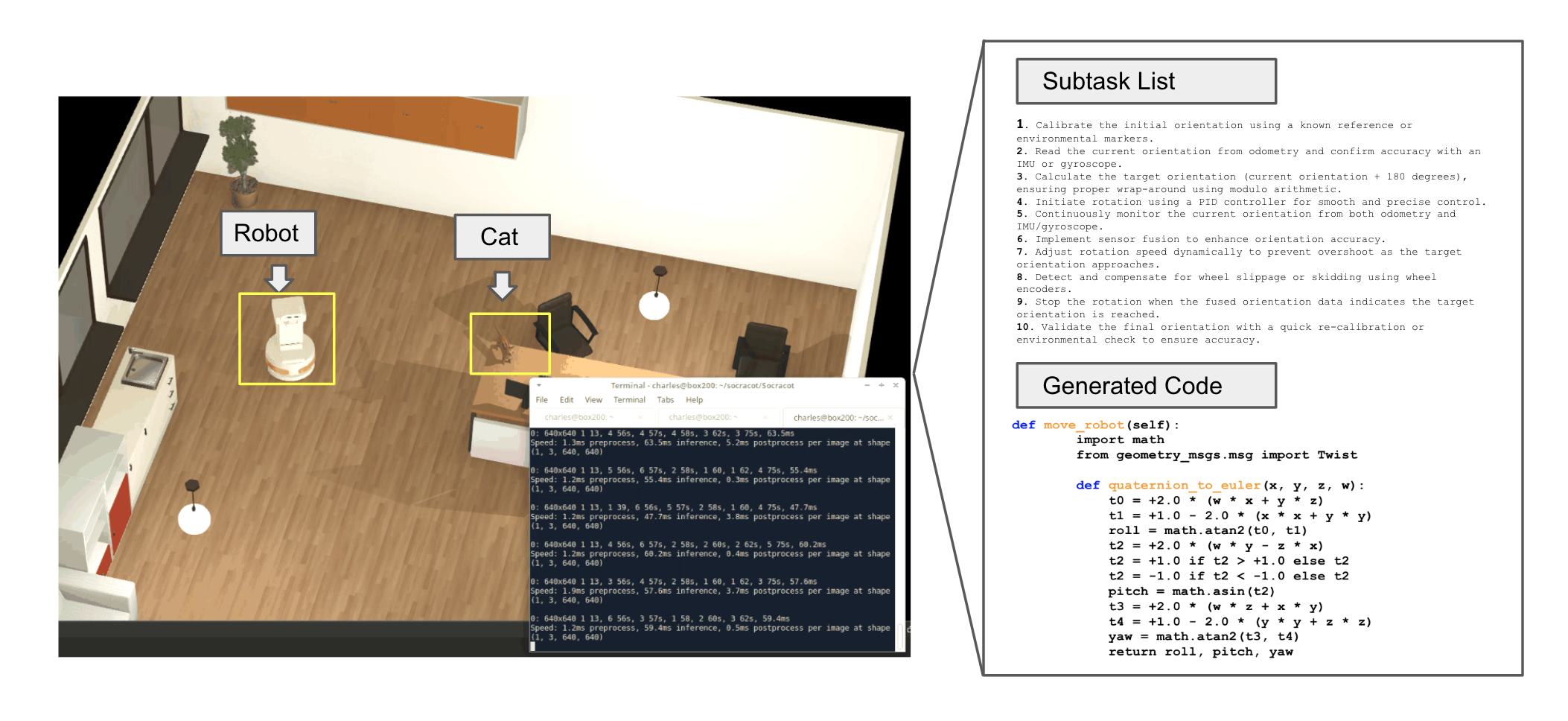}
    \caption{Simulation environment demonstrating object orientation, final subtask list from LLM1 and LLM2 debators, and code sequence generated by LLM3 (moderator).}
    \label{fig:demo}
\end{figure}

\section{Discussion}
\subsection{Limitations}
We would like to address shortcomings in our evaluation method, with emphasis on the need for more rigorous performance metrics for quality of subtask list, quality
of code, and task completion rate (e.g. formulate scores for length and computational efficiency). It is also of interest to develop a robust prompt engineering method to ensure reasoning strategies are fully optimized, and enable a true ablation test for the effectiveness of the Socratic method. It is possible that inconclusive findings regarding the effectiveness of SocraCoT arose from suboptimal prompting. For example, we observed that the LLM struggles with spatial perception and state estimation. This could possibly stem from the fact the LLM only keeps faint knowledge of ROS2 Odometry messages, and needs to output code to unpack its value so that linear and angular positional information can be useful. 
It is also rumored that GPT-4 has undergone architectural changes in its backend by employing the mixture-of-experts \cite{moe} method for increased computational efficiency. As a tradeoff, the response from GPT-4, even with the same hyperparameter and prompt, is no longer consistent; it depends on which gating network receives the input query and which network the input is routed to. Such behavior of GPT-4 could be detrimental to code generation tasks, as it requires consistent performance to generate working and robust code. A slight deviation could lead to broken or non-working code, which consists of over 90\% percent of failures in our experiment.

\subsection{Prompt strategy}
There are several improvements that could be applied to our prompt strategy. The most obvious improvement is the incorporation of more domain knowledge in the prompt. We could add more functions in the control interface script, regarding localization and perception tasks, and by utilizing the set of working functions, the LLM can focus more on handling high-level planning and decision tasks instead of spending tokens in dealing with lower-level tasks. A RAG (retrieval-augmented generation)\cite{RAG} module can also be added for LLM to lookup robotics related knowledge stored in a vector database and offload some domain-specific tasks. In addition, we could increase the number of rounds for debate, while having an LLM agent evaluate their solutions. The evaluation scores from multiple rounds of debate could be used for cherry-picking the final solution that is applied for task completion.

\subsection{Component level improvement} 
Another possible improvement is to refactor certain components within the robotics tasks. One method is to take advantage of Wasserstein distance and mutual information as in \cite{evince}. Despite it is challenging to compute Wasserstein distance for generated code, we propose a new framework through EVINCE framework to aid with robotics localization and perception, thus mitigating LLM's struggle in spatial awareness. Here we propose the adaptation of EVINCE algorithm on object localization as described in alg. \ref{alg:one}. Furthermore, we hereby propose EVINCE-LoC, a modified EVINCE method, hoping to tackle the challenges of object localization and state estimation.

\begin{algorithm}
\caption{EVINCE-LoC}\label{alg:one}
\begin{algorithmic}[1]
\State \textbf{Collect Input}: Actively explore the environment, collect a trajectory sample data $\mathcal{S} = \{ \textbf{Object}_i, \textbf{Odometry}_i\}$, with each frame at timestamp $i$ including odometer and detected objects. Stop when you are have discovered the objects of interest;
\State \textbf{Input}: Sample data $\mathcal{S}$, voxelized space $\mathcal{C}$; Two equally competent LLMs: $\textbf{LLM}_A$ and $\textbf{LLM}_B$;
\State \textbf{Output}: $P_f$, final probability distribution over voxelized space $\mathcal{C}$;
\State \textbf{Variables}: $t$: debate round; $R = \emptyset$ aggregated arguments;
$P_A^{(t)}, P_B^{(t)}$: prediction distributions of $\textbf{LLM}_A$ and $\textbf{LLM}_B$ on $\mathcal{C}$ of round $t$; $R_A^{(t)}$ and $R_B^{(t)}$: supporting data within sensor frames, $\Delta = 90\%$: debate contentiousness, initialize to high to foster adversary between LLMs(opposite corners of a room); $p$: prompt = "Predict positional probability distribution of $<$\textbf{OBJECT}$>$ on voxelized space $\mathcal{C}$ with $\mathcal{S}$ and $R$ at contentiousness $\Delta$";
\State \textbf{Functions}: \textbf{CRIT} \cite{EY2023}, Critical Reading Inquisitive Template for evaluating argument quality;
\textbf{ARA} \cite{Guo24}, Algorithmic Robust Aggregation for optimal prediction aggregation
\State \textbf{Initial Predictions} $t = 0$:
LLMs generate their predictions in probability distributions with supporting sensor frames:
$
    (P_A^{(t = 0)}, R_A^{(t)}) = \textbf{LLM}_A(\mathcal{S}, p), (P_B^{(t = 0)}, R_B^{(t)}) = \textbf{LLM}_B(\mathcal{S}, p)
$
\State \textbf{Debate Iterations}: 
\begin{enumerate}
    \item \textbf{Update Predictions}: \par
    Calculate the confidence-based weights using the inverse of entropy:
    $\alpha = 1 / (H(P_A^{(t)} + 1), \beta = 1 / (H(P_B^{(t)} + 1)$ \par
    Use the blending mechanism to update predictions:
    $P^{'(t)}_A = \alpha P^{(t)}_A + (1 - \alpha) P_B^{(t)}$, \par 
    $P^{'(t)}_B = \beta P^{(t)}_B + (1 - \beta) P_A^{(t)}$
    \item \textbf{LLMs Generate New Predictions}: Both LLMs use accumulated $R = R \cup R_A^{(t)} \cup R_B^{(t)}$ \par
    $
    (P_A^{(t+1)}, R_A^{(t+1)}) = \textbf{LLM}_A(P_B^{'(t)}, R, p), $ \par
    $(P_B^{(t+ 1)}, R_B^{(t+1)}) = \textbf{LLM}_B(P_A^{'(t)}, R, p)
    $
    \item \textbf{Exit Condition Check with Wasserstein distance}: \par
        \textbf{If} \textbf{WD}($P_A^{(t+1)}, P_B^{(t+1)}$) $< \epsilon$ \textbf{EXIT}; $t = t + 1$, $\Delta = \Delta \times 80\%$.
\end{enumerate}
\State \textbf{Final Decision}: Weighted prediction by quality scores of the evaluator e.g., \textbf{CRIT} \par
$P_f = \Omega_A P_A^{t+1} + \Omega_B P_B^{(t+1)} / \Omega_A + \Omega_B$
\State \textbf{Output Location}: $P(\text{OBJECT}) = \arg \max_{k} P_f(k)$, in which $k$ denotes the index of voxels in the environment. 
\end{algorithmic}
\end{algorithm}

\section{Conclusion}
We wish to further explore the effectiveness of SocraCoT through prompt engineering and modulation of task difficulty. It is of interest to see whether Socratic reasoning, when combined with a series of refinement options, enables the robot to succeed in highly complex and or dynamic tasks.

\vspace{12pt}


\begin{thebibliography}{00}

\bibitem{lamport94}
Sai Vemprala, Rogerio Bonatti, Arthur Bucker, and Ashish Kapoor.
\emph{ChatGPT for Robotics: Design Principles and Model Abilities, https://arxiv.org/abs/2306.17582}.
\textbf{arXiv}, 2023.


\bibitem{Yao3}
Yao Mu, Qinglong Zhang, Mengkang Hu, Wenhai Wang, Mingyu Ding, Jun Jin, Bin Wang, Jifeng Dai, Yu Qiao, and Ping Luo.
\emph{EmbodiedGPT: Vision-Language Pre-Training via Embodied Chain of Thoughts, https://proceedings.neurips.cc/paper\_files/paper}.
\textbf{NeurIPS}, 2023.

\bibitem{Zihan2023}
Zihan Yu, Liang He, Zhen Wu, Xinyu Dai, and Jiajun Chen.
\emph{Towards Better Chain-of-Thought Prompting Strategies:
A Survey, https://arxiv.org/pdf/2310.04959}.
\textbf{arXiv}, 2023.

\bibitem{Xue2023}
Xue Jiang, Yihong Dong, Lecheng Wang, Zheng Fang, Qiwei Shang, Ge Li , Zhi Jin, and Wenpin Jiao.
\emph{Self-planning Code Generation with Large
Language Models, https://arxiv.org/pdf/2303.06689}.
\textbf{arXiv}, 2023.

\bibitem{Igor2023}
Igor Mordatch, Yilun Du, Antonio Torralba, Shuang Li, Joshua Tanenbaum.
\emph{Improving Factuality and Reasoning in Language Models through Multiagent Debate, https://arxiv.org/abs/2204.00598}.
\textbf{arXiv}, 2023.

\bibitem{Andy2023}
Andy Zheng, Maria Attarian, Brian Ichter, Stefan Welker, Federico Tombari, Aveek Purohit, Johnny Lee, Vincent Vanhoucke, Pete Florence.
\emph{Socratic Models: Composing Zero-Shot
Multimodal Reasoning with Language, https://arxiv.org/abs/2305.14325}.
\textbf{arXiv}, 2023.

\bibitem{EY2023}
Edward Y. Chang
\emph{Prompting Large Language Models With the Socratic Method, https://arxiv.org/abs/2303.08769}.
\textbf{arXiv}, 2023.

\bibitem{evince}
Edward Y. Chang
\emph{EVINCE (Entropy Variation and Information Competence), Stanford InfoLab Technical Report}
June 6th, 2024

\bibitem{moe}
Robert A. Jacobs, Michael I. Jordan, Steven J. Nowlan,
and Geoffrey E. Hinton. 
\emph{Adaptive Mixtures of
Local Experts. Neural Computation, 3(1):79–87.}
1991

\bibitem{Guo24}
Yongkang Guo, Jason D. Hartline, Zhihuan Huang,
Yuqing Kong, Anant Shah, and Fang-Yi Yu
\emph{Algorithmic robust forecast aggregation. Preprint,
arXiv:2401.17743.}
2024

\bibitem{RAG}
Yunfan Gao, Yun Xiong, Xinyu Gao, Kangxiang Jia, Jinliu Pan, Yuxi Bi, Yi Dai, Jiawei Sun, Meng Wang, Haofen Wang
\emph{Retrieval-Augmented Generation for Large Language Models: A Survey. https://arxiv.org/abs/2312.10997}
2023

\end{thebibliography}
\end{document}